\documentclass[11pt]{article}

\usepackage[margin=1in]{geometry}
\usepackage{amsmath,amssymb,amsfonts}
\usepackage{graphicx}
\usepackage{booktabs}
\usepackage{hyperref}
\usepackage{url}
\usepackage[numbers,sort&compress]{natbib}
\usepackage{xcolor}
\usepackage{enumitem}
\usepackage{algorithm}
\usepackage{algpseudocode}
\usepackage{caption}
\usepackage{subcaption}
\usepackage[breakable]{tcolorbox}
\usepackage{tikz}
\usepackage{multirow}
\usepackage{amsthm}
\usepackage{pifont}
\usepackage{float}

\newtheorem{definition}{Definition}
\newtheorem{proposition}{Proposition}

\hypersetup{
    colorlinks=true,
    linkcolor=blue!70!black,
    citecolor=blue!70!black,
    urlcolor=blue!70!black
}

\title{\textbf{Nurture-First Agent Development: Building Domain-Expert AI Agents \\Through Conversational Knowledge Crystallization}}

\author{
    Linghao Zhang\thanks{Corresponding author. GitHub: \url{https://github.com/ZLHad}. The author is an active user of OpenClaw and Claude Code, whose architectures motivated the NFD framework presented in this paper.} \\[4pt]
    \textit{Jiangsu Key Laboratory of Wireless Communications} \\
    \textit{Nanjing University of Posts and Telecommunications (NJUPT)} \\
    \textit{Nanjing, Jiangsu, China} \\[2pt]
    \texttt{zhangczssx@gmail.com}
}

\date{}

\begin{document}
\maketitle

% ============================================================
% Abstract
% ============================================================
\begin{abstract}
The emergence of large language model (LLM)-based agent frameworks has shifted the primary challenge in building domain-expert AI agents from raw capability to effective encoding of domain expertise. Two dominant paradigms---\emph{code-first} development, which embeds expertise in deterministic pipelines, and \emph{prompt-first} development, which captures expertise in static system prompts---both treat agent construction as a discrete engineering phase preceding deployment. We argue that this sequential assumption creates a fundamental mismatch with the nature of domain expertise, which is substantially tacit, deeply personal, and continuously evolving. We propose \textbf{Nurture-First Development (NFD)}, a paradigm in which agents are initialized with minimal scaffolding and progressively \emph{grown} through structured conversational interaction with domain practitioners. The central mechanism is the \emph{Knowledge Crystallization Cycle}, whereby fragmented knowledge embedded in operational dialogue is periodically consolidated into structured, reusable knowledge assets. We formalize NFD through: (1)~a \emph{Three-Layer Cognitive Architecture} organizing agent knowledge by volatility and personalization degree; (2)~the \emph{Knowledge Crystallization Cycle} with formal definitions of crystallization operations and efficiency metrics; and (3)~an operational framework comprising a \emph{Dual-Workspace Pattern} and \emph{Spiral Development Model}. We illustrate the paradigm through a detailed case study on building a financial research agent for U.S.\ equity analysis and discuss the conditions, limitations, and broader implications of NFD for human-agent co-evolution.
\end{abstract}


% ============================================================
% 1. Introduction
% ============================================================
\section{Introduction}
\label{sec:introduction}

The rapid maturation of large language model (LLM)-based agent frameworks---exemplified by AutoGPT~\cite{autogpt2023}, MetaGPT~\cite{hong2024metagpt}, AutoGen~\cite{wu2023autogen}, and personal agent infrastructures like OpenClaw~\cite{openclaw2025} and Claude Code~\cite{anthropic2025claude}---has shifted the bottleneck in AI agent development from raw capability to knowledge encoding~\cite{xi2023rise}. The central question is no longer \emph{``Can an agent perform task X?''} but \emph{``How do we encode the domain expertise that makes an agent's outputs genuinely valuable?''} We term this gap the \emph{configuration gap}: the disparity between the general capabilities of foundation models and the domain-specific expertise required for agents to produce outputs a practitioner would trust for consequential decisions.

Two dominant paradigms address this challenge.

\textbf{The Code-First Paradigm.} Developers build domain agents by writing deterministic pipelines, API integrations, and rule-based logic~\cite{chase2022langchain, liu2022llamaindex}. This paradigm yields reliable, reproducible agents but struggles to capture nuanced, judgment-heavy aspects of expert practice. A code-first medical triage agent can reliably classify symptoms but cannot replicate the interpretive judgment of a clinician who recognizes that a symptom constellation means something different in the context of a patient's full history. The paradigm's deeper limitation is temporal: code captures the developer's understanding at the time of writing, and updating requires expensive engineering cycles.

\textbf{The Prompt-First Paradigm.} Users encode expertise through elaborate system prompts, persona definitions, and few-shot demonstrations~\cite{bai2022constitutional, khattab2024dspy}. This approach is more accessible but faces a scaling problem: as domain complexity grows, the prompt expands until it exceeds context-window limits or degrades inference quality~\cite{wei2022chain}. Moreover, prompts are static snapshots with no built-in mechanism for learning from operational experience. Prompt optimization tools like DSPy~\cite{khattab2024dspy} can tune parameters, but the fundamental artifact remains a fixed text.

Both paradigms share a critical assumption: that agent development is a \emph{discrete phase that precedes deployment}. We argue that this assumption creates a structural mismatch with how domain expertise is acquired and maintained.

As Polanyi~\cite{polanyi1966tacit} and Nonaka and Takeuchi~\cite{nonaka1995knowledge} observed, domain expertise is substantially \emph{tacit}---practitioners know more than they can articulate at any given moment. Expert decision-making is shaped by pattern recognition, contextual judgment, and accumulated case memory that resists comprehensive upfront formalization~\cite{dreyfus1986mind, collins1989cognitive}. Moreover, expertise \emph{evolves}: practitioners continuously update their mental models based on new experiences. A static encoding begins degrading in value from the moment of its creation.

We propose a third paradigm: \textbf{Nurture-First Development (NFD)}. In this paradigm, the agent is not ``built'' and then ``deployed''---it is \emph{born} with minimal scaffolding and then \emph{raised} through sustained interaction with its user. Development and deployment are concurrent, interleaved processes. The agent's domain expertise grows organically through daily use, accumulates in its memory system as fragmented experiential data, and is periodically consolidated into structured knowledge assets through deliberate \emph{crystallization} processes.

This paradigm draws direct inspiration from modern agent infrastructures. OpenClaw~\cite{openclaw2025} exemplifies the enabling architecture: workspace-first design with structured identity files, modular skills loaded on demand, persistent memory with semantic search, and community-contributed skill ecosystems. Claude Code~\cite{anthropic2025claude} demonstrates how agentic tools with persistent project memory can co-evolve with developers through sustained interaction. These platforms implicitly support the nurture-first pattern---our contribution is to formally articulate and systematize it as a development methodology.

The contributions of this paper are:
\begin{enumerate}[leftmargin=*]
    \item A formal characterization of the Nurture-First Development paradigm and its principled distinction from code-first and prompt-first approaches (Section~\ref{sec:paradigm}).
    \item A \emph{Three-Layer Cognitive Architecture} that organizes agent knowledge by volatility and personalization degree, enabling clear separation between scaffolded structure and nurtured content (Section~\ref{sec:architecture}).
    \item The \emph{Knowledge Crystallization Cycle} as the core developmental mechanism, with formal definitions of crystallization operations, efficiency metrics, and an algorithmic specification (Section~\ref{sec:crystallization}).
    \item A \emph{Dual-Workspace Pattern} and \emph{Spiral Development Model} that operationalize NFD into a practical development workflow (Section~\ref{sec:operational}).
    \item An illustrative case study applying NFD to the construction of a financial research agent for U.S.\ equity analysis (Section~\ref{sec:case_study}).
\end{enumerate}

\begin{figure}[H]
    \centering
    \includegraphics[width=0.78\textwidth]{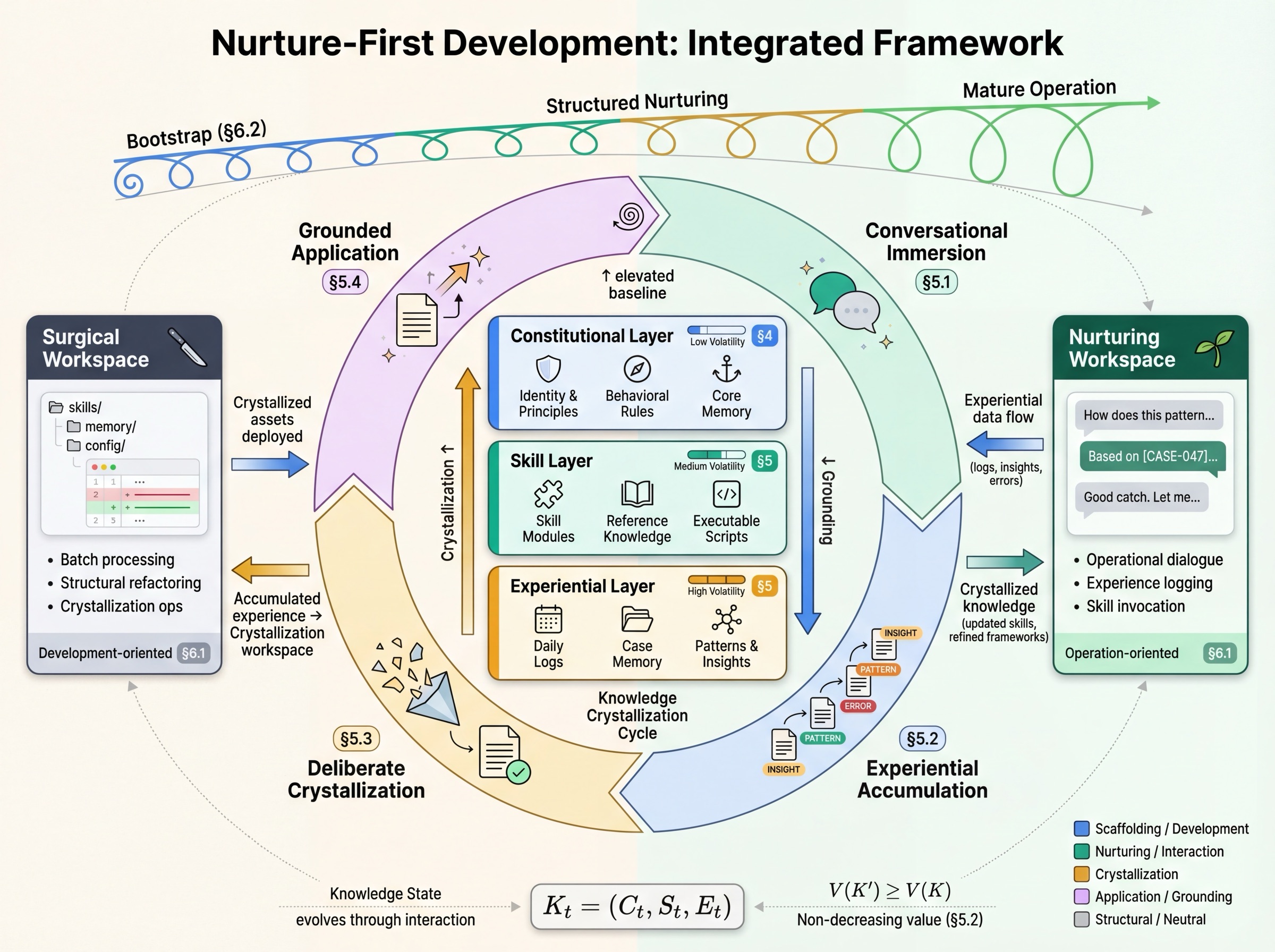}
    \caption{\textbf{Integrated overview of the Nurture-First Development framework.} The diagram unifies the four core contributions: \ding{192}~the \emph{Three-Layer Cognitive Architecture} (center stack) organizing knowledge by volatility---Constitutional (blue, low), Skill (teal, medium), and Experiential (amber, high); \ding{193}~the \emph{Knowledge Crystallization Cycle} (surrounding ring) with four phases---Conversational Immersion, Experiential Accumulation, Deliberate Crystallization, and Grounded Application; \ding{194}~the \emph{Dual-Workspace Pattern} (left: Surgical Workspace for crystallization; right: Nurturing Workspace for operational dialogue); and \ding{195}~the \emph{Spiral Development Model} (outer expanding trajectory) through which these components interact across successive development revolutions. Arrows trace the flow of knowledge: tacit expertise enters through conversation (right), accumulates as experiential records (bottom), is crystallized into structured skill references (left), and grounds future interactions at progressively higher baselines (top). Section numbers indicate where each component is formally developed.}
    \label{fig:nfd_overview}
\end{figure}

\noindent Figure~\ref{fig:nfd_overview} provides an integrated overview of the NFD framework, illustrating how these contributions form a coherent paradigm. The Three-Layer Cognitive Architecture (center) organizes agent knowledge by volatility; the Knowledge Crystallization Cycle (surrounding ring) drives progressive knowledge consolidation; the Dual-Workspace Pattern (flanking panels) separates developmental and operational concerns; and the Spiral Development Model (outer trajectory) governs the macro-level lifecycle. The remainder of this paper develops each component in turn.

% FIGURE: NFD Framework Overview (full-width, placed at top of next page)

% ============================================================
% 2. Related Work
% ============================================================
\section{Related Work}
\label{sec:related}

\textbf{Agent Frameworks and Architectures.}
The agent landscape has evolved from monolithic systems to modular, composable architectures. LangChain~\cite{chase2022langchain} and LlamaIndex~\cite{liu2022llamaindex} popularized pipeline-based approaches; ReAct~\cite{yao2023react} introduced the reasoning-and-acting loop now foundational to modern designs. AutoGPT~\cite{autogpt2023} pioneered autonomous goal decomposition but exposed the fragility of fully autonomous agents. CrewAI~\cite{crewai2024}, MetaGPT~\cite{hong2024metagpt}, and AutoGen~\cite{wu2023autogen} explored multi-agent collaboration. Sumers et al.~\cite{sumers2024cognitive} proposed cognitive architectures drawing on classical AI, and comprehensive surveys have charted the broader landscape~\cite{xi2023rise}. OpenClaw~\cite{openclaw2025} introduced workspace-first architecture with persistent memory, modular skills, and structured identity files---features that implicitly enable the nurture-first pattern we formalize here.

\textbf{Knowledge Engineering and Elicitation.}
The challenge of encoding expert knowledge into computational systems has a long history in knowledge engineering~\cite{schreiber2000knowledge}. Traditional approaches involved structured interviews and protocol analysis to extract and formalize expert knowledge into rule-based systems~\cite{alavi2001knowledge}. Nonaka and Takeuchi's SECI model~\cite{nonaka1995knowledge} describes organizational knowledge transformation through four modes: Socialization (tacit to tacit), Externalization (tacit to explicit), Combination (explicit to explicit), and Internalization (explicit to tacit). Our Knowledge Crystallization Cycle can be understood as an operationalization of the \emph{Externalization} mode within a human-agent interaction context, where the agent's memory system serves as the medium for tacit-to-explicit conversion. The Dreyfus model of skill acquisition~\cite{dreyfus1986mind}---which identifies five stages from novice to expert---further informs our understanding of how expertise develops through progressive stages, a trajectory that NFD mirrors in the agent context.

\textbf{Interactive Machine Teaching and Human-AI Collaboration.}
Interactive Machine Teaching (IMT)~\cite{simard2017teaching} explores how humans can teach machine learning systems through natural interaction rather than formal training data. Amershi et al.~\cite{amershi2019guidelines} established guidelines for human-AI interaction design, emphasizing the importance of supporting user-driven customization and learning from user behavior. Our approach extends IMT into the agent domain, encompassing not just concept transfer but the progressive co-construction of a shared cognitive framework between human and agent. Kolb's experiential learning theory~\cite{kolb1984experiential} and Lave and Wenger's situated learning framework~\cite{lave1991situated} provide theoretical grounding for our emphasis on learning through situated practice rather than abstract instruction.

\textbf{Memory-Augmented and Self-Improving Agents.}
Retrieval-Augmented Generation (RAG)~\cite{lewis2020rag} established the paradigm of augmenting LLM outputs with retrieved context. MemGPT~\cite{packer2023memgpt} introduced OS-inspired memory management with explicit tiers, while Generative Agents~\cite{park2023generative} demonstrated emergent behavior through experience-based memory. Reflexion~\cite{shinn2023reflexion} showed improvement through verbal self-reflection, Voyager~\cite{wang2023voyager} demonstrated skill acquisition through a growing library, and Madaan et al.~\cite{madaan2023selfrefine} introduced iterative self-refinement. ExpeL~\cite{zhao2024expel} demonstrated that agents can extract reusable insights from task experience without parameter updates, and recent work on self-evolving agents~\cite{gao2025evolver} and lifelong learning roadmaps~\cite{zhao2025lifelong} has begun formalizing continuous improvement through experience. These systems provide infrastructure upon which NFD builds, yet none frames the process as a holistic development methodology centered on the human-agent co-construction of domain expertise.

\textbf{Prompt Engineering and Optimization.}
DSPy~\cite{khattab2024dspy} formalized prompt optimization as a programmatic process, while Constitutional AI~\cite{bai2022constitutional} introduced the idea of encoding behavioral principles as persistent constraints. RLHF~\cite{ouyang2022training} demonstrated how human feedback can align model behavior with user preferences. The prompt-first paradigm draws on these traditions. However, prompt engineering treats the prompt as a \emph{static artifact} to be optimized, whereas NFD treats the agent's knowledge base as a \emph{living system} that grows through interaction. The distinction is analogous to the difference between compiling a textbook and mentoring an apprentice.

% ============================================================
% 3. The NFD Paradigm
% ============================================================
\section{The Nurture-First Development Paradigm}
\label{sec:paradigm}

\subsection{Core Propositions}
\label{sec:propositions}

We define Nurture-First Development through three foundational propositions:

\begin{tcolorbox}[colback=gray!5, colframe=gray!60, title={\small\textbf{Proposition 1: Development-Deployment Fusion}}]
\small Agent development and deployment are not sequential phases but \emph{concurrent, interleaved} processes. The agent becomes operational before its knowledge base is complete, and its knowledge base continues to grow during operational use.
\end{tcolorbox}

\begin{tcolorbox}[colback=gray!5, colframe=gray!60, title={\small\textbf{Proposition 2: Conversational Knowledge Acquisition}}]
\small The primary channel for encoding domain expertise into the agent is \emph{natural language conversation} between the user and the agent during normal operational use, rather than upfront formal specification.
\end{tcolorbox}

\begin{tcolorbox}[colback=gray!5, colframe=gray!60, title={\small\textbf{Proposition 3: Crystallization as Development}}]
\small The core ``development'' activity in NFD is not writing code or crafting prompts, but periodically \emph{crystallizing} accumulated conversational fragments into structured, reusable knowledge assets.
\end{tcolorbox}

These propositions collectively redefine the development lifecycle. Rather than a linear pipeline from specification to deployment, NFD operates as a continuous spiral where each interaction simultaneously serves operational and developmental purposes. The metaphor of ``nurturing'' is deliberate: like mentoring an apprentice, the process is incremental, bidirectional, and yields results that emerge organically from sustained interaction.

\subsection{Paradigm Comparison}
\label{sec:comparison}

% FIGURE 1: Three-Paradigm Comparison
\begin{figure}[H]
    \centering
    \includegraphics[width=0.95\textwidth]{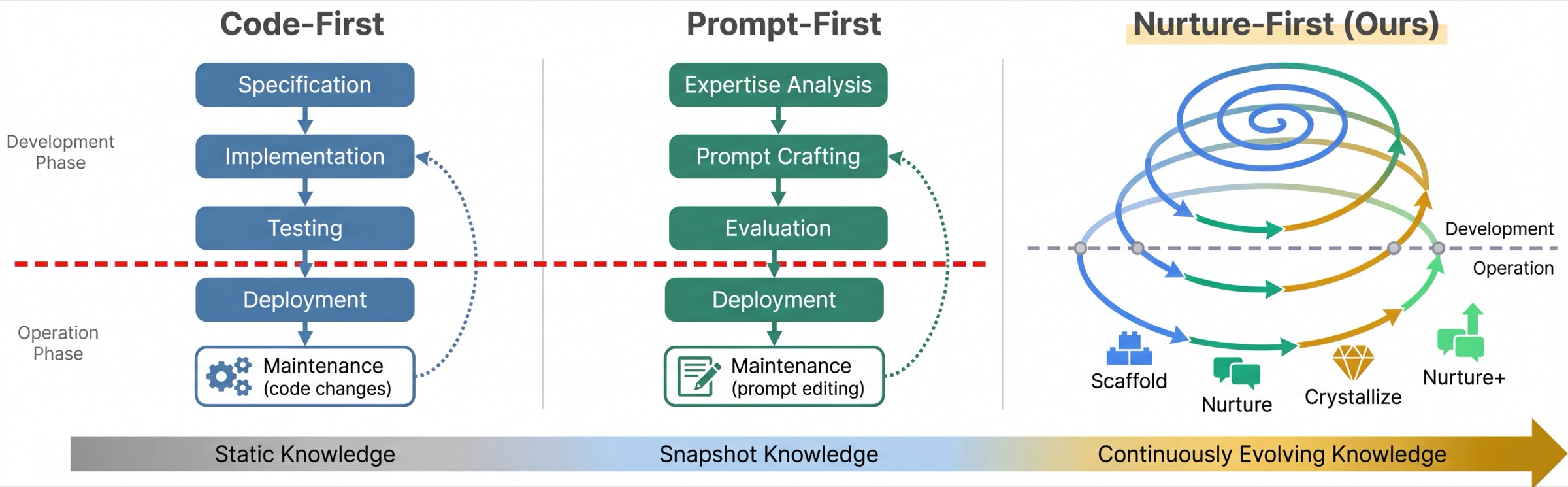}
    \caption{\textbf{Three paradigms of agent development.} Code-First and Prompt-First follow a linear develop-then-deploy lifecycle with a hard boundary between construction and operation. Nurture-First dissolves this boundary: a spiral of scaffolding, nurturing, and crystallization phases interleaves development with deployment, enabling continuous knowledge growth throughout the agent's operational lifetime.}
    \label{fig:paradigm_comparison}
\end{figure}

Table~\ref{tab:paradigm_comparison} summarizes the key differences across the three paradigms along ten critical dimensions. Figure~\ref{fig:paradigm_comparison} illustrates the structural difference in development lifecycles.

\begin{table}[t]
\centering
\caption{Comparative analysis of agent development paradigms across ten dimensions.}
\label{tab:paradigm_comparison}
\small
\begin{tabular}{@{}p{3.2cm}p{3.6cm}p{3.6cm}p{4.8cm}@{}}
\toprule
\textbf{Dimension} & \textbf{Code-First} & \textbf{Prompt-First} & \textbf{Nurture-First (NFD)} \\
\midrule
Knowledge encoding & Deterministic code & Static prompt text & Evolving memory files \\
Development phase & Before deployment & Before deployment & Continuous, concurrent \\
Expertise capture & Explicit logic only & Declarative snapshot & Tacit + explicit, evolving \\
Update mechanism & Code change + redeploy & Prompt editing & Conversation + crystallization \\
Personalization & Parameterization & Prompt customization & Deep experiential adaptation \\
Primary developer & Software engineer & Prompt engineer & Domain practitioner (user) \\
Time to first value & Weeks--months & Hours--days & Minutes (scaffold), growing \\
Knowledge asset & Codebase & Prompt library & Memory corpus + skill refs \\
Scalability ceiling & Engineering capacity & Context window & Memory search quality \\
Transferability & High (code is portable) & Medium (prompts portable) & Low--Med (personalized) \\
\bottomrule
\end{tabular}
\end{table}

A critical distinction lies in \emph{who develops the agent}. In code-first, a software engineer maintains logic; in prompt-first, a prompt engineer crafts prompts; in nurture-first, the \emph{domain practitioner} is the primary developer---their daily interaction constitutes the development process. This democratizes agent development for practitioners with deep domain expertise but limited engineering skills.

The \emph{scalability ceiling} also differs fundamentally. Code-first agents are limited by engineering capacity; prompt-first agents by context window size. NFD agents are bounded by memory search quality: vast experiential knowledge can be accumulated, but practical utility depends on how effectively retrieval surfaces relevant memories at inference time.

\subsection{Applicability Conditions}
\label{sec:applicability}

NFD is not universally superior to existing paradigms. It is most appropriate when the following conditions hold:

\begin{enumerate}[leftmargin=*]
    \item \textbf{Domain expertise is substantially tacit.} The practitioner cannot fully articulate their decision-making process upfront but can recognize and explain their reasoning in context when confronted with specific situations~\cite{polanyi1966tacit}.
    \item \textbf{Expertise is highly personal.} Different practitioners in the same domain have legitimately different approaches, and the agent's value derives from alignment with a \emph{specific} practitioner's framework rather than encoding a universal best practice.
    \item \textbf{Expertise evolves continuously.} The domain environment changes frequently enough that any static encoding requires regular updates, making the maintenance cost of code-first or prompt-first agents prohibitively high.
    \item \textbf{The interaction pattern is conversational.} The user interacts with the agent through dialogue, providing natural opportunities for knowledge transfer during operational use.
    \item \textbf{Experiential pattern recognition is valuable.} The agent's ability to recall and reference the user's past experiences, decisions, and outcomes is a significant source of analytical value.
\end{enumerate}

Domains satisfying these criteria include professional advisory work (legal, medical, financial), academic research, creative practice, strategic planning, and any skilled practice where ``judgment'' is the differentiating factor. Domains where expertise is fully formalizable and static (e.g., tax form processing) are better served by code-first approaches.

% ============================================================
% 4. Three-Layer Cognitive Architecture
% ============================================================
\section{Three-Layer Cognitive Architecture}
\label{sec:architecture}

A central challenge in NFD is organizing the agent's growing knowledge base so that different knowledge types are stored, accessed, and updated through mechanisms matched to their characteristics. We propose a Three-Layer Cognitive Architecture that partitions knowledge along two dimensions: \textbf{volatility} (how frequently the knowledge changes) and \textbf{personalization degree} (how specific the knowledge is to an individual user). This design draws on hierarchical memory systems in cognitive science~\cite{kolb1984experiential} and tiered memory architectures in recent agent systems~\cite{packer2023memgpt, sumers2024cognitive}.

% FIGURE 2: Three-Layer Architecture
\begin{figure}[H]
    \centering
    \includegraphics[width=0.78\textwidth]{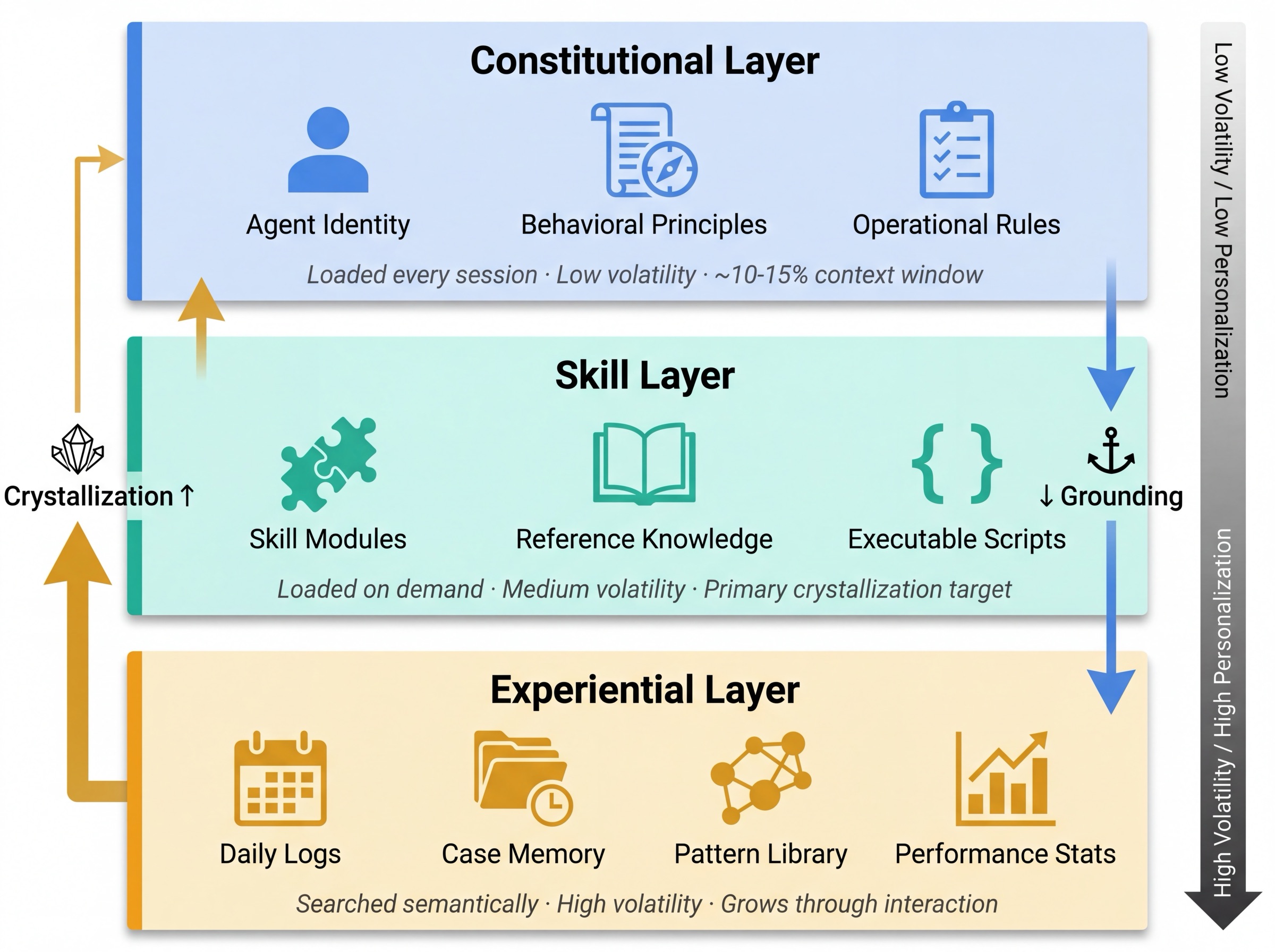}
    \caption{\textbf{Three-Layer Cognitive Architecture.} Knowledge is organized into three layers by volatility and personalization degree. The Constitutional Layer (low volatility, loaded every session) contains identity and principles. The Skill Layer (medium volatility, loaded on demand) contains structured domain knowledge. The Experiential Layer (high volatility, searched semantically) contains accumulated operational experience. Upward \emph{crystallization} arrows represent knowledge consolidation; downward \emph{grounding} arrows represent interpretive application.}
    \label{fig:architecture}
\end{figure}

\subsection{Constitutional Layer}

The Constitutional Layer contains the agent's foundational identity, behavioral principles, and operational rules. In modern agent architectures such as OpenClaw~\cite{openclaw2025}, this corresponds to bootstrap files loaded into the system prompt on every session---persona definitions (\texttt{SOUL.md}), behavioral boundaries, operational rules (\texttt{AGENTS.md}), user profile (\texttt{USER.md}), and core memory summaries (\texttt{MEMORY.md}).

\textbf{Characteristics:} (a)~Loaded every session as part of the system prompt; (b)~low volatility, updated only when fundamental beliefs or principles change; (c)~low-to-medium personalization; (d)~acts as an implicit constraint on all agent behavior, including how other layers are interpreted; (e)~size-sensitive, as it consumes context window on every interaction, typically occupying 10--15\% of available context.

\textbf{Design Principle:} The Constitutional Layer should contain \emph{indices and principles}, not detailed knowledge. A constitutional directive such as ``Always apply the risk assessment framework before making recommendations'' is appropriate; the full specification of that framework belongs in a Skill Layer reference file. This indexing strategy ensures that the Constitutional Layer remains concise while providing navigational pointers to deeper knowledge.

\subsection{Skill Layer}

The Skill Layer contains modular, task-specific capabilities packaged as reusable units. Each skill encapsulates: (i)~instructional prompts defining how to perform a specific task, (ii)~reference knowledge files containing domain knowledge needed for the task, and (iii)~optionally, executable scripts for deterministic data processing. In OpenClaw's architecture, these are organized as skill folders containing \texttt{SKILL.md}, \texttt{references/}, and \texttt{scripts/} directories.

\textbf{Characteristics:} (a)~Loaded on demand when the agent judges relevance to the current task; (b)~medium volatility, updated periodically as domain understanding deepens through crystallization; (c)~medium personalization---the skill's task logic is generalizable, but reference knowledge may encode a practitioner's specific analytical framework; (d)~the primary container for crystallized knowledge assets; (e)~no inherent size constraint beyond practical file length limits.

\textbf{Design Principle:} Skills should follow the Single Responsibility Principle---each handling one well-defined task. Complex workflows emerge from the agent's reasoning about which skills to invoke, not from hard-coded dependencies. Inter-skill coordination occurs through shared memory files. Community-contributed skill ecosystems~\cite{openclaw2025} enable rapid capability expansion while maintaining modularity.

\subsection{Experiential Layer}

The Experiential Layer contains the agent's accumulated operational experience: daily interaction logs, case memories, identified behavioral patterns, performance statistics, and all knowledge arising from use rather than specification. This corresponds to persistent memory directories with dated entries (\texttt{memory/YYYY-MM-DD.md}) and specialized memory files (error patterns, case libraries).

\textbf{Characteristics:} (a)~Accessed via semantic search (vector-based retrieval) or direct file read; (b)~high volatility, with new entries added daily; (c)~high personalization, inherently unique to each user-agent pair; (d)~the raw material from which crystallization produces structured Skill Layer knowledge; (e)~grows indefinitely, requiring periodic curation to maintain signal-to-noise ratio.

\textbf{Design Principle:} The Experiential Layer should be optimized for write-heavy, search-friendly operation. Daily logs should be append-only; a separate curation process synthesizes durable insights upward into higher layers. Temporal decay mechanisms~\cite{packer2023memgpt} naturally handle relevance ranking of old versus new experiences.

\subsection{Cross-Layer Information Flow}

The three layers participate in a dynamic, bidirectional information flow (Figure~\ref{fig:architecture}):

\textbf{Downward flow (Grounding):} Constitutional principles and Skill knowledge provide the interpretive framework for new experiences. The agent reads relevant Skill references constrained by Constitutional principles; the experiential record is stored with this interpretive context attached, enriching future retrievability.

\textbf{Upward flow (Crystallization):} Accumulated experiential data is periodically consolidated into structured knowledge updating the Skill Layer or, less frequently, the Constitutional Layer. This is the core developmental mechanism of NFD (Section~\ref{sec:crystallization}).

\textbf{Lateral flow (Memory-mediated coordination):} Skills coordinate through shared memory files rather than direct invocation. This loose coupling avoids context explosion from loading multiple Skills simultaneously while enabling emergent cross-skill reasoning.

% ============================================================
% 5. Knowledge Crystallization Cycle
% ============================================================
\section{The Knowledge Crystallization Cycle}
\label{sec:crystallization}

The Knowledge Crystallization Cycle (KCC) is the core developmental mechanism of NFD---the process by which fragmented, contextual knowledge embedded in conversational interactions is progressively transformed into structured, reusable, and transferable knowledge assets. Drawing on Nonaka and Takeuchi's concept of knowledge externalization~\cite{nonaka1995knowledge}, the KCC operationalizes tacit-to-explicit conversion within a human-agent interaction context, with the agent's memory system serving as the medium for transformation.

\subsection{Four Phases}

\textbf{Phase 1: Conversational Immersion.}
The user and agent engage in normal operational interaction---analyzing situations, making decisions, reflecting on outcomes. Knowledge transfer occurs implicitly through dialogue rather than explicit ``teaching'' sessions. When a user explains ``This reminds me of last quarter---the same pattern of early consensus followed by sudden reversal,'' they transfer complex, context-rich pattern recognition that resists upfront formalization. The agent gains access to the user's \emph{reasoning process}, not just conclusions. Immersion is most effective when the agent possesses enough baseline knowledge to be a meaningful conversational partner.

\textbf{Phase 2: Experiential Accumulation.}
Every interaction generates experiential data logged in the agent's memory system. We identify six categories of experiential knowledge:
\begin{enumerate}[leftmargin=*, label=(\roman*)]
    \item \emph{Operational Records}---decisions made, actions taken, outcomes observed.
    \item \emph{Reasoning Traces}---explanations of decision logic, including assumptions and alternatives considered.
    \item \emph{Pattern Observations}---regularities noticed across multiple experiences (``every time X happens, Y tends to follow'').
    \item \emph{Error Records}---mistakes and their analysis, including what went wrong and the corrective principle.
    \item \emph{Contextual Annotations}---environmental metadata providing interpretive context.
    \item \emph{Insight Fragments}---standalone principles articulated during conversation (``I've realized that...'' or ``the key thing about this domain is...'').
\end{enumerate}

Not all experiential data requires crystallization. Insight fragments may be immediately promotable to higher layers. Error records typically require accumulation before patterns become visible. Operational records contribute primarily to aggregate statistics. The semi-structured nature of these categories, facilitated by consistent tagging (e.g., \texttt{[DECISION]}, \texttt{[INSIGHT]}, \texttt{[ERROR]}), enables efficient extraction during crystallization.

% FIGURE: Knowledge Crystallization Cycle (moved here for better page flow)
\begin{figure}[H]
    \centering
    \includegraphics[width=0.65\textwidth]{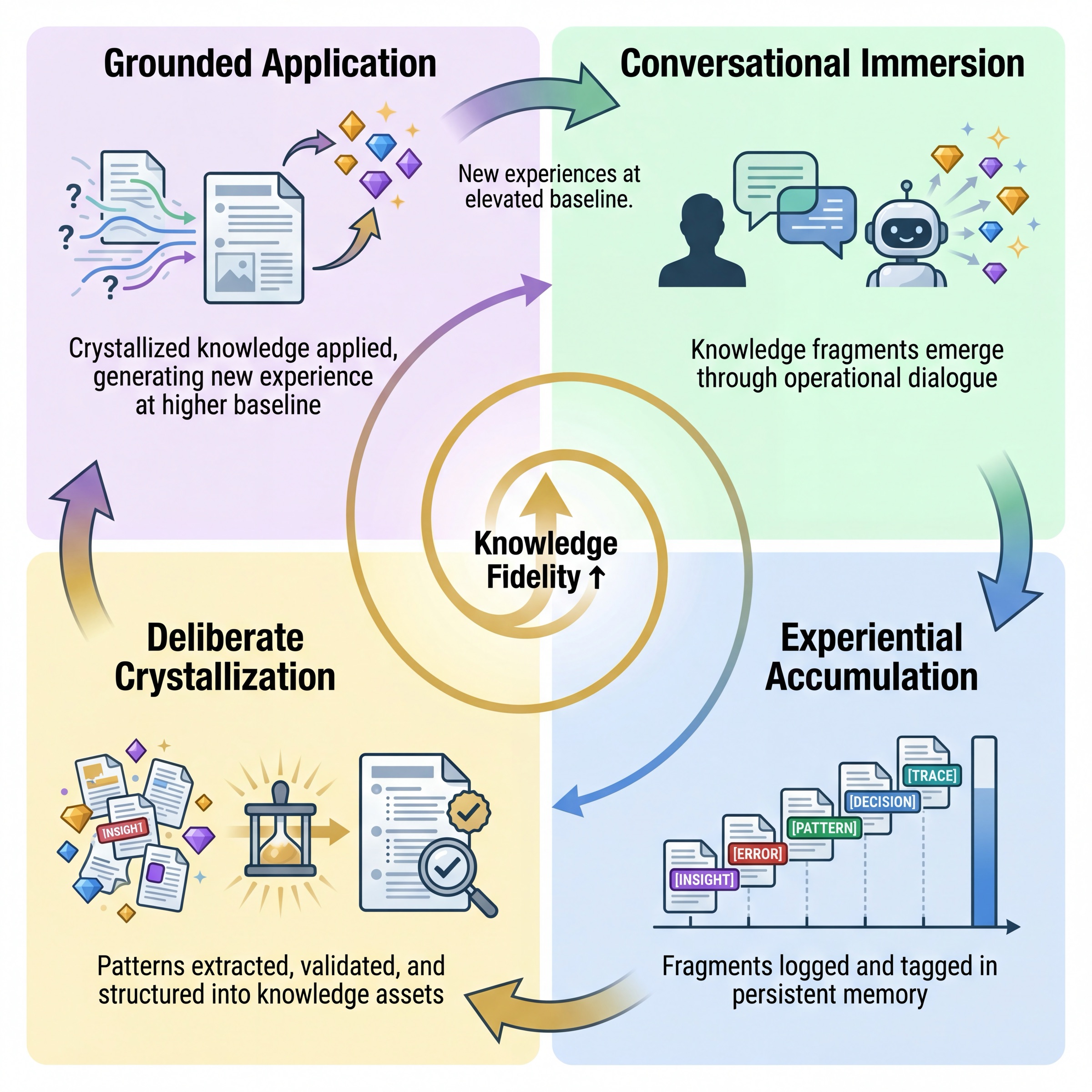}
    \caption{\textbf{The Knowledge Crystallization Cycle.} Four phases form an ascending spiral: (1)~Conversational Immersion generates knowledge fragments through operational dialogue; (2)~Experiential Accumulation logs and tags these fragments in persistent memory; (3)~Deliberate Crystallization consolidates patterns into structured knowledge assets; (4)~Grounded Application deploys crystallized knowledge in practice, generating new experiences at a higher baseline. Each revolution raises the agent's knowledge fidelity.}
    \label{fig:crystallization}
\end{figure}

\textbf{Phase 3: Deliberate Crystallization.}
Unlike the organic first two phases, crystallization is a deliberate, periodic process that typically requires the surgical workspace (Section~\ref{sec:operational}). It involves five sub-operations: (a)~\emph{Pattern Extraction}---mining the experiential corpus for recurring patterns through both automated analysis and human review; (b)~\emph{Knowledge Structuring}---organizing extracted patterns into Skill Layer formats (reference documents, decision frameworks, case libraries); (c)~\emph{De-contextualization}---removing situation-specific details to produce generalizable principles; (d)~\emph{Validation}---checking crystallized knowledge against the full experiential corpus to confirm patterns are genuinely supported; and (e)~\emph{Integration}---writing validated assets into appropriate layers with version tracking.

\textbf{Phase 4: Grounded Application.}
Crystallized knowledge enters active service in subsequent interactions, improving immediate performance while generating new experiential data that may confirm, refine, or \emph{challenge} crystallized patterns. This feedback loop is essential: crystallization is not one-way distillation but a \emph{hypothesis-generating process}. Crystallized patterns are hypotheses that must be continuously tested against new experience, with contradictions triggering re-crystallization.

\subsection{Formal Model}
\label{sec:formal_model}

We formalize the key constructs of the Knowledge Crystallization Cycle.

\begin{definition}[Agent Knowledge State]
An agent's knowledge state at time $t$ is a tuple $K_t = (C_t, S_t, E_t)$, where $C_t$ is the Constitutional Layer content, $S_t = \{s_1, s_2, \ldots, s_n\}$ is the set of Skill Layer reference contents, and $E_t = \{e_1, e_2, \ldots, e_m\}$ is the set of Experiential Layer entries.
\end{definition}

\begin{definition}[Experiential Accumulation]
After a conversational interaction $I_t = (q_t, r_t, K_t)$ at time $t$ (where $q_t$ is the user input and $r_t$ the agent response), the experiential layer is updated:
\begin{equation}
    E_{t+1} = E_t \cup \{\delta(I_t)\}
\end{equation}
where $\delta: \mathcal{I} \to \mathcal{E}$ is the experience extraction function producing a structured experiential record from the raw interaction.
\end{definition}

\begin{definition}[Knowledge Crystallization]
A crystallization operation at time $\tau$ is a function $\kappa$ that transforms the current knowledge state:
\begin{equation}
    K_{\tau}' = \kappa(K_{\tau}) = (C_{\tau}', S_{\tau}', E_{\tau}')
\end{equation}
where typically $|E_{\tau}'| \leq |E_{\tau}|$ (experiential entries are consolidated) and $\mathcal{H}(S_{\tau}') \geq \mathcal{H}(S_{\tau})$ (information content of the Skill Layer increases or remains equal).
\end{definition}

We model the value of an agent's knowledge state as:
\begin{equation}
    V(K_t) = \alpha \cdot \mathrm{Breadth}(E_t) + \beta \cdot \mathrm{Structure}(S_t) + \gamma \cdot \mathrm{Align}(C_t, U)
    \label{eq:value}
\end{equation}
where $\mathrm{Breadth}(E_t)$ measures the coverage and diversity of experiential data, $\mathrm{Structure}(S_t)$ measures the quality and completeness of crystallized skill knowledge, $\mathrm{Align}(C_t, U)$ measures the degree to which constitutional principles match the user $U$'s actual values and preferences, and $\alpha, \beta, \gamma$ are domain-dependent weights. The dominant term shifts over the agent's lifecycle: in early nurturing, $\alpha$ dominates (new experiences are highly valuable); after initial crystallization, $\beta$ dominates (structured knowledge enables qualitatively better reasoning); in mature agents, $\gamma$ differentiates agents of comparable capability.

We define crystallization efficiency as:
\begin{equation}
    \eta(\kappa, E) = \frac{\Delta \mathrm{Structure}(S)}{|E_{\text{consumed}}|}
    \label{eq:efficiency}
\end{equation}
Crystallization efficiency depends on three factors: (i)~\emph{diversity} of accumulated experiences---a corpus with varied situations crystallizes better than one with repetitive data; (ii)~\emph{annotation quality}---experiences with explicit reasoning traces crystallize more efficiently than bare operational records; and (iii)~\emph{pattern density} inherent in the domain---domains with strong regularities yield higher efficiency than those dominated by idiosyncratic situations.

\begin{proposition}[Non-decreasing Value]
Under the assumption that crystallization operations are validated (i.e., only genuinely supported patterns are promoted to the Skill Layer), the value function is non-decreasing across crystallization cycles:
\begin{equation}
    V(K_{\tau}') \geq V(K_{\tau}) \quad \forall\, \tau \text{ where } K_{\tau}' = \kappa(K_{\tau})
    \label{eq:nondecreasing}
\end{equation}
This follows from the definition of crystallization: $\mathcal{H}(S_{\tau}') \geq \mathcal{H}(S_{\tau})$ ensures $\mathrm{Structure}(S)$ does not decrease, while validated crystallization preserves the experiential coverage captured by $\mathrm{Breadth}(E)$, since consolidated entries retain their informational content in structured form.
\end{proposition}

This property distinguishes NFD from paradigms where knowledge updates may introduce regressions. The human validation step in crystallization (Algorithm~\ref{alg:crystallization}, line~3) serves as the critical safeguard ensuring monotonic improvement.

\subsection{Crystallization Process}
\label{sec:crystallization_algorithm}

Algorithm~\ref{alg:crystallization} formalizes the crystallization process as a sequence of operations performed in the surgical workspace.

\begin{algorithm}[t]
\caption{Knowledge Crystallization Process}
\label{alg:crystallization}
\small
\begin{algorithmic}[1]
\Require Knowledge state $K_\tau = (C_\tau, S_\tau, E_\tau)$, scope $\theta$
\Ensure Updated knowledge state $K_\tau'$
\State $\mathcal{D} \gets \textsc{ScopeFilter}(E_\tau, \theta)$ \Comment{Select relevant entries}
\State $\mathcal{P} \gets \textsc{ExtractPatterns}(\mathcal{D})$ \Comment{Automated pattern detection}
\State $\mathcal{P}^* \gets \textsc{HumanReview}(\mathcal{P})$ \Comment{User validates patterns}
\State $\mathcal{A} \gets \emptyset$ \Comment{Knowledge assets to produce}
\For{each validated pattern $p \in \mathcal{P}^*$}
    \State $k \gets \textsc{Structure}(p, S_\tau)$ \Comment{Format for Skill Layer}
    \State $k \gets \textsc{Decontextualize}(k)$ \Comment{Generalize}
    \If{$\textsc{Validate}(k, E_\tau)$} \Comment{Check against full corpus}
        \State $\mathcal{A} \gets \mathcal{A} \cup \{k\}$
    \EndIf
\EndFor
\State $S_\tau' \gets \textsc{Integrate}(S_\tau, \mathcal{A})$ \Comment{Update Skill Layer}
\State $E_\tau' \gets \textsc{Archive}(E_\tau, \mathcal{P}^*)$ \Comment{Consolidate entries}
\State $C_\tau' \gets \textsc{UpdatePrinciples}(C_\tau, \mathcal{A})$ \Comment{If warranted}
\State \Return $K_\tau' = (C_\tau', S_\tau', E_\tau')$
\end{algorithmic}
\end{algorithm}

The algorithm highlights the \emph{human-in-the-loop} nature of crystallization: automated pattern extraction proposes candidates, but user validation (line~3) ensures that only meaningful patterns are promoted. The validation step (line~8) provides an additional safeguard by checking that crystallized knowledge is genuinely supported by the full experiential corpus, not just the scoped subset.

\subsection{Crystallization Triggers}

We identify three triggering modes: (a)~\emph{Scheduled crystallization}---performed at regular intervals (weekly, monthly), suitable for domains with steady experiential accumulation; (b)~\emph{Threshold-triggered crystallization}---initiated when the volume of un-crystallized experiential data exceeds a configurable threshold or when performance metrics indicate behavioral drift; and (c)~\emph{Event-triggered crystallization}---initiated after significant domain events (e.g., regime changes, project phase transitions, regulatory updates) that may require updating crystallized knowledge. In practice, most implementations combine scheduled and event-triggered crystallization, with the user exercising final judgment on timing.

% ============================================================
% 6. Operational Framework
% ============================================================
\section{Operational Framework}
\label{sec:operational}

\subsection{The Dual-Workspace Pattern}

\begin{figure}[H]
    \centering
    \includegraphics[width=0.95\textwidth]{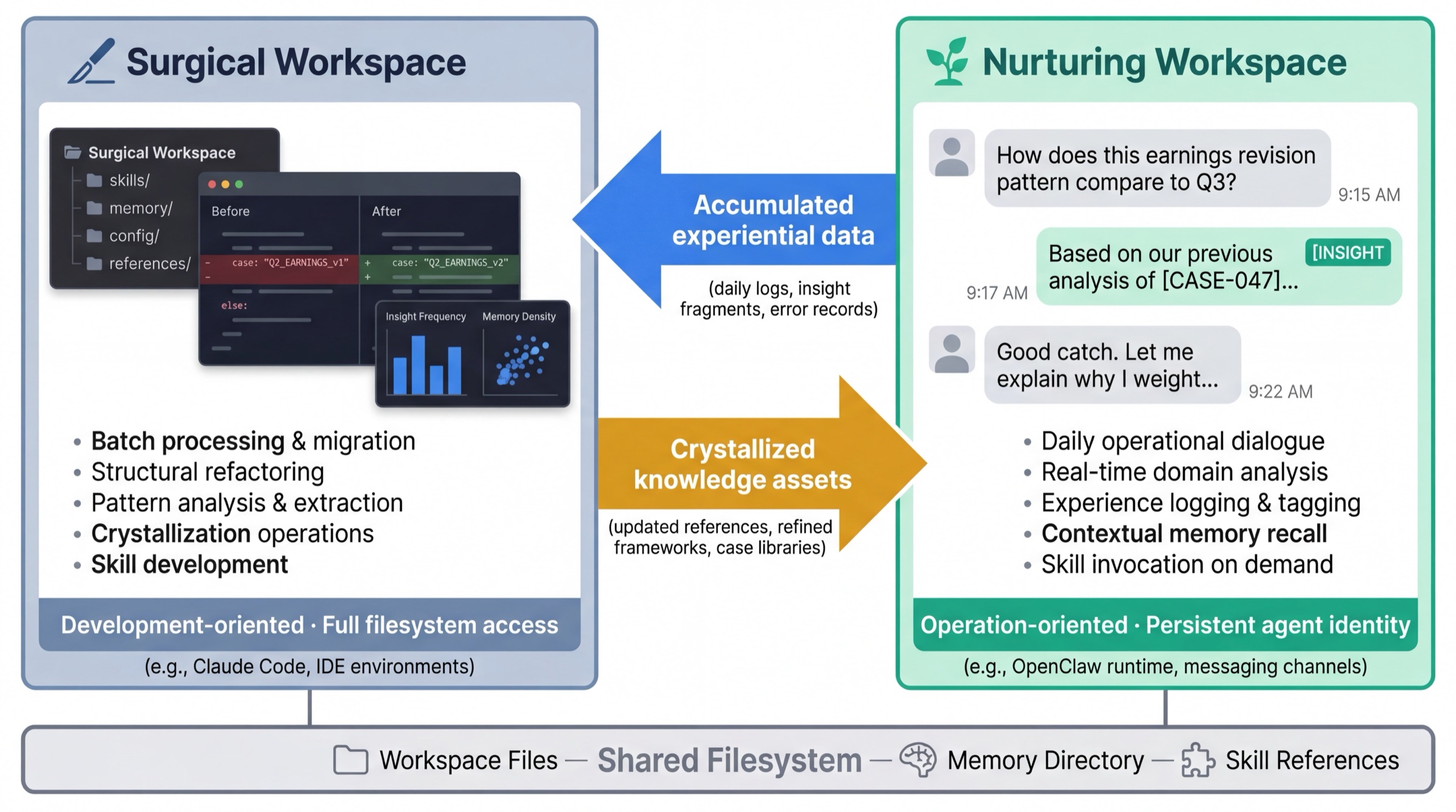}
    \caption{\textbf{The Dual-Workspace Pattern.} The \emph{Surgical Workspace} (left) provides full filesystem access for batch processing, crystallization, and skill development. The \emph{Nurturing Workspace} (right) is the agent's runtime environment for daily conversational interaction. Both operate on shared state---the agent's file system---enabling seamless knowledge transfer between development and operational modes.}
    \label{fig:dual_workspace}
\end{figure}

NFD operationally requires two distinct interaction environments that serve complementary purposes (Figure~\ref{fig:dual_workspace}).

% FIGURE 4: Dual-Workspace Pattern

The \textbf{Surgical Workspace} is a development environment with full programmatic access to the agent's file system---the workspace directory, memory files, skill definitions, and configuration. It handles operations that are structural, batch-oriented, or analytically complex: initial scaffolding of the workspace, bulk processing of historical data into memory format, crystallization operations requiring analysis of large experiential corpora, skill development and refactoring, and statistical analysis of agent performance. The surgical workspace treats the agent's knowledge base as a \emph{data structure to be analyzed and optimized}. Agentic coding tools such as Claude Code~\cite{anthropic2025claude}, with their persistent project memory and file manipulation capabilities, are well-suited for this role.

The \textbf{Nurturing Workspace} is the agent's runtime environment---the conversational channel through which the user and agent interact in daily operational use. The agent maintains its persistent identity, accesses memory through semantic retrieval, and invokes skills as needed. This is where conversational immersion and experiential accumulation occur naturally. Platforms like OpenClaw~\cite{openclaw2025}, with their persistent memory, skill invocation infrastructure, and multi-channel communication, provide the foundation for this workspace.

The key insight is \emph{separation of concerns}: the surgical workspace handles structural development (scaffolding, crystallization, refactoring), while the nurturing workspace handles conversational growth (immersion, accumulation). Both operate on shared state---the agent's file system---enabling seamless knowledge transfer. This separation reflects a fundamental difference in interaction modality: surgical operations demand precision and batch-processing capability; nurturing operations demand continuity, identity persistence, and conversational fluency.

\subsection{The Spiral Development Model}

\begin{figure}[H]
    \centering
    \includegraphics[width=0.78\textwidth]{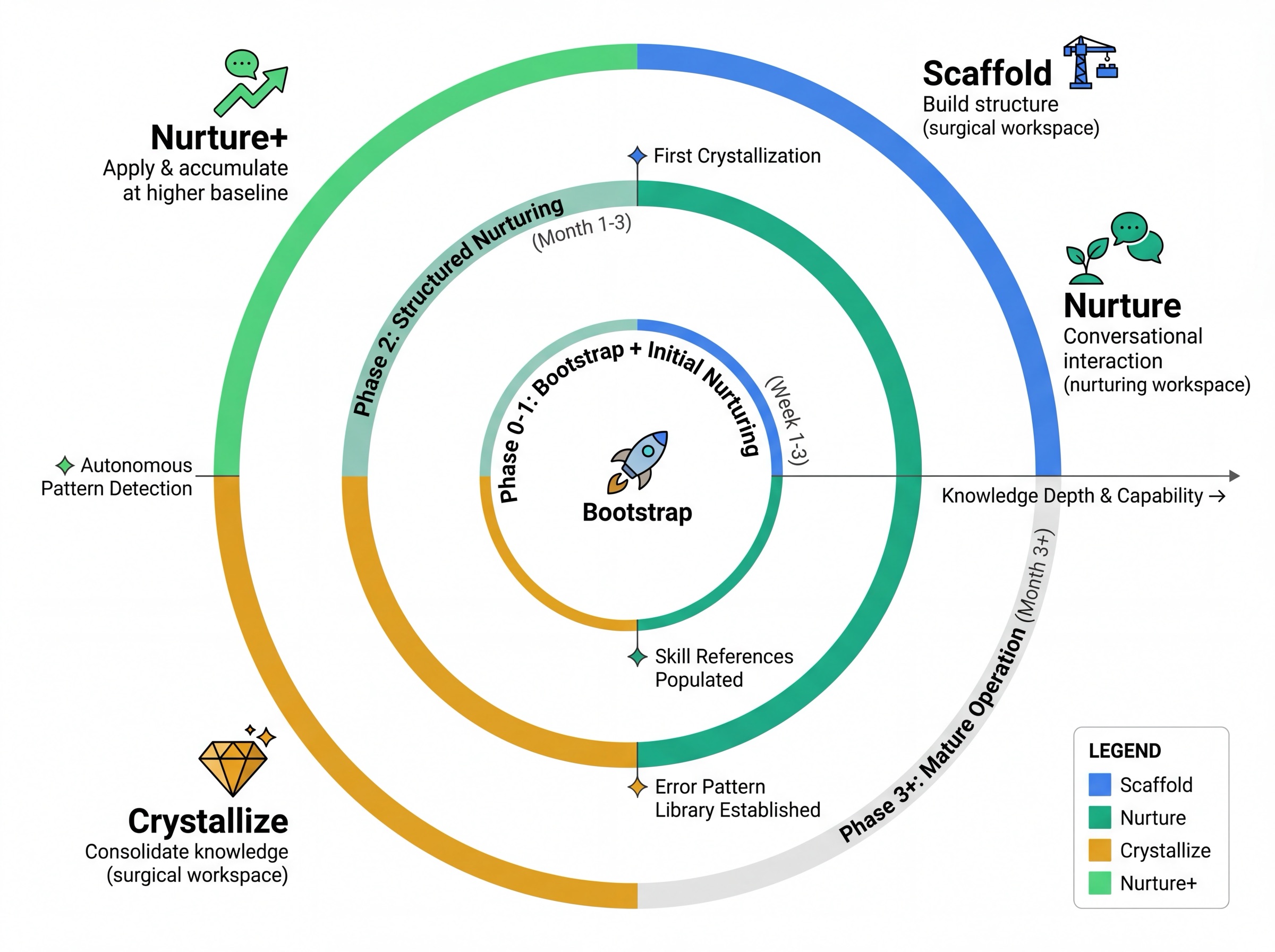}
    \caption{\textbf{The Spiral Development Model.} An expanding spiral alternates between four quadrants: Scaffold (surgical workspace builds structure), Nurture (conversational interaction accumulates experience), Crystallize (surgical workspace consolidates knowledge), and Nurture$^+$ (agent applies crystallized knowledge at a higher baseline). Inner revolutions correspond to early phases; outer revolutions to mature operation. Radial expansion represents growing knowledge depth.}
    \label{fig:spiral}
\end{figure}

NFD follows a spiral development trajectory~\cite{boehm1988spiral} where scaffolding and nurturing phases alternate, connected by crystallization checkpoints (Figure~\ref{fig:spiral}).

% FIGURE 5: Spiral Development Model

\textbf{Phase~0: Bootstrap} (1--3~days). The surgical workspace creates minimal viable scaffolding: directory structure, skeleton bootstrap files with initial persona and principles, and basic skill definitions. The goal is \emph{bootability}, not completeness. If historical data exists, a \textbf{Historical Data Migration} accelerates early phases by converting records into memory format, annotating them with category tags, and running preliminary pattern extraction---compressing months of experiential accumulation into days of surgical processing.

\textbf{Phase~1: Initial Nurturing} (1--3~weeks). The user begins daily interaction, establishing Constitutional Layer elements through dialogue (beliefs, principles, risk attitudes) while generating the first wave of experiential data through normal operational use.

\textbf{Crystallization Checkpoint~1.} The surgical workspace performs the first crystallization: updating memory summaries with emergent principles, creating initial skill reference files from discussed domain knowledge, and identifying structural issues. This checkpoint often reveals discrepancies between the user's self-reported framework and actual practice---a discovery that is itself valuable.

\textbf{Phase~2: Structured Nurturing} (1--3~months). With crystallized knowledge now available in skill references, the agent becomes a more capable conversational partner. Interactions shift from foundational knowledge transfer to operational collaboration---analyzing real situations together, applying frameworks, generating and discussing recommendations. The experiential corpus grows in both volume and quality.

\textbf{Crystallization Checkpoint~2.} A deeper crystallization: extracting behavioral patterns from the accumulated operational record, creating specialized memory files for recurring knowledge structures (case libraries, error pattern databases), and refining skill reference files based on how knowledge has been applied in practice.

\textbf{Phase~3+: Mature Operation.} With a substantial experiential base and well-crystallized skill references, crystallization becomes routine maintenance. Focus shifts to edge cases, novel situations, and continuous refinement. The agent may begin proactively proposing crystallization candidates based on self-detected patterns.

% ============================================================
% 7. Case Study: Financial Research Agent
% ============================================================
\section{Case Study: Financial Research Agent}
\label{sec:case_study}

To ground the NFD methodology, we present a case study of building a financial research agent designed to serve as an equity research partner for U.S.\ public-market analysis. This domain satisfies all five applicability conditions: equity analysis expertise is substantially tacit (analysts develop intuitive pattern recognition), highly personal (different analysts maintain legitimately different frameworks), continuously evolving (market regimes shift), conversational (analyst work involves extensive discussion and interpretation), and pattern-recognition-intensive.

% FIGURE 6: Case Study Overview
\begin{figure}[H]
    \centering
    \includegraphics[width=0.95\textwidth]{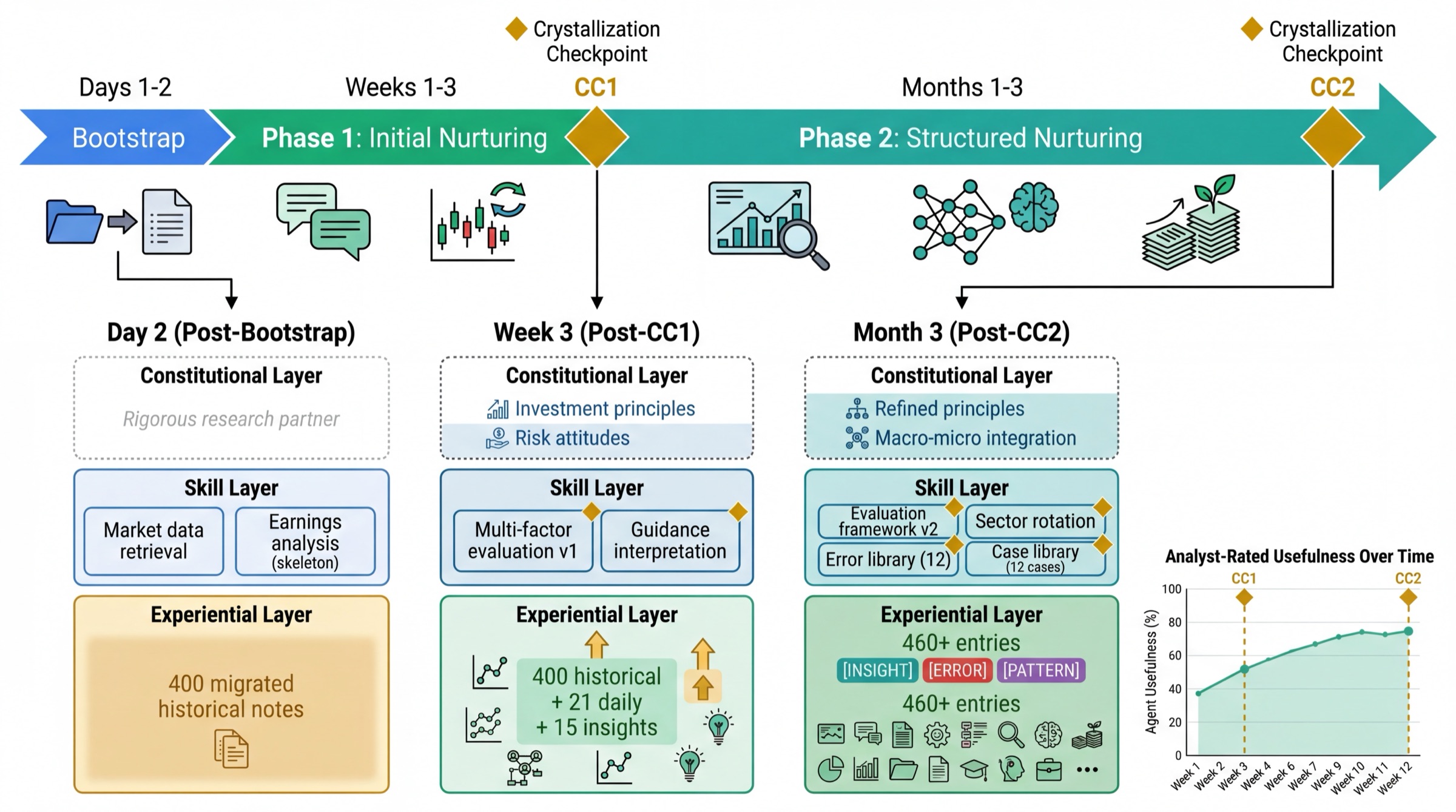}
    \caption{\textbf{NFD applied to financial research agent development.} The timeline shows the spiral development trajectory: from initial scaffolding (market data skills, skeleton analytical framework) through nurturing phases (daily market discussions, earnings analysis) to crystallization checkpoints (extracting evaluation frameworks, risk heuristics). Below: progressive population of the three cognitive layers. Inset: growth in agent usefulness over 12 weeks.}
    \label{fig:case_study}
\end{figure}

\textbf{Context.} An equity research analyst with over five years of experience in U.S.\ equity markets sought to build an AI agent that functions as a research and decision-support partner. The analyst possessed a corpus of historical research notes ($\sim$400 entries spanning 18~months) and a partially articulated multi-factor evaluation framework applied intuitively but never fully documented.

\textbf{Bootstrap Phase (Days~1--2).} Using the surgical workspace, the initial scaffolding was constructed: workspace directory structure, a persona file defining the agent as a ``rigorous, evidence-based research partner,'' and basic skills for market data retrieval, earnings analysis, and sector comparison. Historical research notes were migrated into the memory format via bulk processing. Preliminary pattern extraction identified 10~recurring analytical themes, 6~frequent judgment errors, and 3~distinct strategic approaches.

\textbf{Phase~1: Initial Nurturing (Weeks~1--3).} During three weeks of daily interaction, the analyst and agent engaged in real-time market discussions. Several important knowledge transfers occurred naturally:
\begin{itemize}[leftmargin=*]
    \item The analyst explained their multi-factor evaluation framework across 5~conversations, revealing assessment criteria that had never been documented---including how the weighting of factors shifts depending on the macroeconomic environment.
    \item During earnings season, the agent captured the analyst's nuanced view on how management guidance language correlates with subsequent performance---a form of tacit pattern recognition developed over years of experience.
    \item The analyst corrected 3~misinterpretations the agent had derived from the historical data, each correction generating a detailed reasoning trace in the experiential log.
    \item 15~insight fragments were captured through natural dialogue, including several the analyst described as ``things I knew but had never articulated.''
\end{itemize}

\textbf{Crystallization Checkpoint~1 (End of Week~3).} Crystallization produced: (a)~a structured multi-factor evaluation reference file for the analytical skill, with explicit factor weights and conditional logic, (b)~an updated memory summary reflecting refined investment principles, (c)~an error pattern library with 8~historical and 2~newly identified judgment biases, and (d)~the discovery that one of the 3 ``approaches'' from the historical data was actually two distinct strategies---the analyst confirmed this, noting they had never explicitly distinguished them before.

\textbf{Phase~2: Structured Nurturing (Months~1--3).} With crystallized frameworks available, the agent became a substantially more capable analytical partner. Table~\ref{tab:metrics} summarizes progression metrics across the development lifecycle.

\begin{table}[t]
\centering
\caption{Agent development progression metrics.}
\label{tab:metrics}
\small
\begin{tabular}{@{}lcccc@{}}
\toprule
\textbf{Metric} & \textbf{Wk 1--3} & \textbf{Post-CC1} & \textbf{Wk 9--12} & \textbf{Post-CC2} \\
\midrule
Useful analyses (\%) & 38 & 52 & 71 & 74 \\
Case recalls & 2 & 5 & 12 & 15 \\
Bias flags & 0 & 1 & 4 & 5 \\
Skill refs populated & 2 & 4 & 6 & 8 \\
Error patterns & 6 & 8 & 10 & 12 \\
Daily log entries & 21 & -- & 60+ & -- \\
\bottomrule
\end{tabular}
\end{table}

The agent successfully recalled relevant historical cases in 12~analytical discussions, drawing from the migrated historical corpus. It flagged potential judgment biases on 4~occasions---the analyst credited 2~of these interventions with improving decision quality. A second crystallization produced refined sector analysis frameworks, enriched case libraries with 12 new documented examples, and a newly formalized approach to integrating macroeconomic indicators into individual equity evaluation.

\subsection{Illustrative Nurturing Episodes}

\begin{figure}[H]
    \centering
    \includegraphics[width=0.95\textwidth]{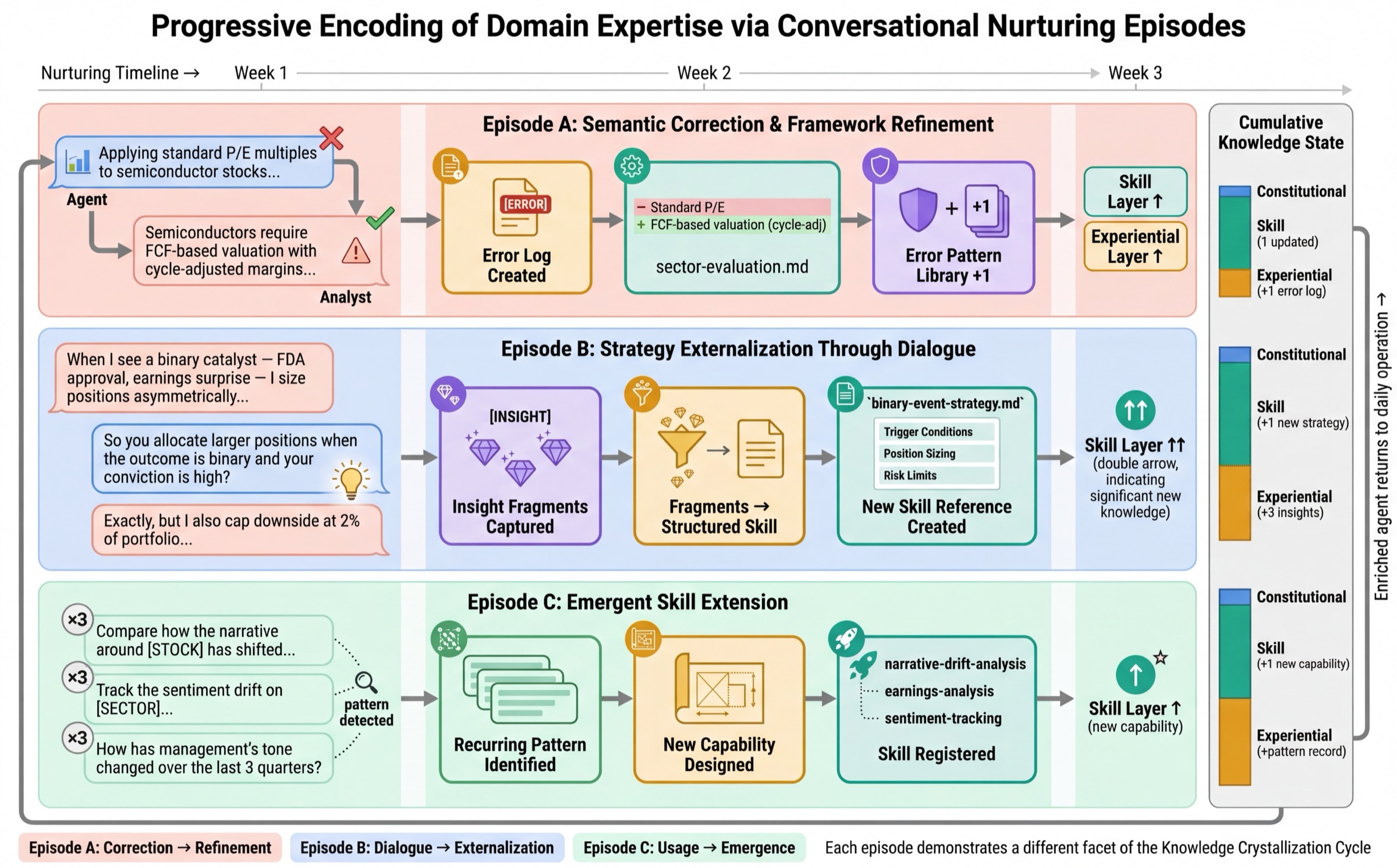}
    \caption{\textbf{Micro-process of conversational nurturing.} Three representative interaction episodes illustrating how domain expertise is progressively encoded through the NFD cycle. \emph{Episode~A}: Semantic correction and framework refinement through error identification. \emph{Episode~B}: Strategy externalization through guided dialogue, where tacit investment logic is made explicit and crystallized into a reusable decision framework. \emph{Episode~C}: Emergent skill extension, where a new analytical capability arises organically from operational need. Arrows indicate knowledge flow between experiential logs, skill references, and constitutional principles.}
    \label{fig:nurturing_detail}
\end{figure}

To concretize how NFD operates at the interaction level, we present three representative episodes that illustrate the core dynamics of conversational nurturing (see Figure~\ref{fig:nurturing_detail} for the micro-process overview).

% FIGURE 7: Detailed Nurturing Interaction Process

\textbf{Episode A: Semantic Correction and Framework Refinement.}
During Week~2, the agent applied its bootstrapped analytical framework to evaluate a semiconductor company and produced a ``strong buy'' recommendation based on high revenue growth and expanding gross margins. The analyst disagreed:

\begin{quote}
\small\emph{``Your analysis is mechanically correct, but you're missing the capital expenditure cycle. In semiconductors, high capex today compresses free cash flow for 2--3 years. You need to weight FCF yield much more heavily for capex-intensive sectors---revenue growth alone is misleading here.''}
\end{quote}

This single correction triggered three knowledge updates: (1)~an experiential log entry tagged \texttt{[ERROR][SECTOR-SPECIFIC]} recording the misapplication and the corrective principle; (2)~a refinement to the multi-factor evaluation skill, adding sector-conditional factor weighting logic (``if capex-intensive sector, increase FCF weight by 1.5$\times$, decrease revenue growth weight by 0.7$\times$''); and (3)~a new entry in the error pattern library documenting ``sector-blind factor application'' as a recurring bias to monitor. At the next crystallization checkpoint, this correction was generalized into a design principle: \emph{all factor-based evaluations must incorporate sector-specific weighting adjustments}---a rule that the analyst had always applied intuitively but had never formalized.

\textbf{Episode B: Strategy Externalization Through Dialogue.}
In Week~5, the analyst asked the agent to analyze a biotech company approaching a key FDA decision. The agent produced a standard risk-reward analysis, but the analyst steered the conversation toward their personal strategy for binary-event situations:

\begin{quote}
\small\emph{``For binary events like FDA decisions, I don't just look at probability-weighted outcomes. I look at what I call `asymmetric conviction'---if the market is pricing in 50/50 odds but I see 70/30 based on the clinical data pattern, the edge isn't in the probability itself but in the \textbf{decay rate of uncertainty}. As trial data accumulates, the market adjusts slowly. I want to be positioned before the inflection point.''}
\end{quote}

This dialogue externalized a sophisticated, deeply personal investment strategy that the analyst had never documented. The agent logged the reasoning trace with tags \texttt{[INSIGHT][STRATEGY][BINARY-EVENT]}. Over subsequent conversations, the analyst elaborated on specific indicators they use to estimate the ``decay rate of uncertainty'' (e.g., insider transaction patterns, conference presentation tone shifts, analogous historical precedents). At Crystallization Checkpoint~2, these fragments were consolidated into a new skill reference file: \texttt{binary-event-strategy.md}, containing the formalized decision framework with entry criteria, position sizing logic, and exit triggers---a structured asset that would have been nearly impossible to produce through a traditional upfront specification process.

\textbf{Episode C: Emergent Skill Extension.}
During Month~2, the analyst repeatedly asked the agent to cross-reference earnings call language across multiple quarters for the same company---comparing how management framing of ``supply chain challenges'' evolved from Q1 to Q3. The agent initially performed this as ad hoc retrieval from experiential logs, but the pattern recurred frequently enough that the analyst remarked:

\begin{quote}
\small\emph{``We keep doing this manually. Can you build a way to systematically track how management narrative evolves quarter over quarter? That drift in language is often more informative than the numbers themselves.''}
\end{quote}

This request triggered a skill extension in the surgical workspace: a new \texttt{narrative-drift-analysis} skill was scaffolded, with an instructional prompt defining the task (extract and compare key narrative themes across sequential earnings transcripts), a reference file encoding the analyst's taxonomy of narrative signals (e.g., ``hedging language increase,'' ``forward guidance specificity,'' ``risk factor emphasis shift''), and a simple script for structured extraction. The skill was initially minimal, but through 8~subsequent uses in the nurturing workspace, the agent accumulated feedback and edge cases that were crystallized into a refined version---demonstrating how \emph{new capabilities emerge organically from operational patterns rather than being specified upfront}.

\medskip

These three episodes illustrate the core dynamics distinguishing NFD from static development approaches: (1)~\emph{corrections refine existing knowledge} rather than being discarded as one-off fixes; (2)~\emph{tacit strategies are externalized incrementally} through contextual dialogue rather than formal elicitation sessions; and (3)~\emph{new capabilities emerge from observed usage patterns} rather than predetermined specifications. Crucially, each episode simultaneously serves an operational purpose (analyzing a real situation) and a developmental purpose (enriching the agent's knowledge base)---the defining characteristic of the nurture-first paradigm.

\subsection{Key Observations and Limitations}

Three notable outcomes emerged from the full case study: First, the most valuable aspect of NFD was the \emph{forced externalization} of reasoning---explaining interpretive judgments to the agent revealed inconsistencies in the analyst's framework that had previously gone unrecognized. Second, the agent's ability to recall and reference the analyst's own historical analyses proved unexpectedly valuable, functioning as an ``institutional memory'' that compensated for natural forgetting. Third, the iterative crystallization process produced methodology documentation more accurate than anything previously written manually, because it was derived from \emph{observed practice} rather than \emph{self-reported theory}.

\textbf{Limitations.} This case study is illustrative rather than definitive. As a single-user deployment without a control group, it demonstrates the \emph{feasibility} of the NFD paradigm but does not constitute a rigorous empirical evaluation. The ``usefulness'' metric reflects subjective practitioner assessment rather than objective performance measurement. Controlled studies with multiple practitioners, standardized evaluation criteria, and comparison against code-first and prompt-first baselines are needed to establish the paradigm's quantitative advantages.

% ============================================================
% 8. Discussion
% ============================================================
\section{Discussion}
\label{sec:discussion}

\subsection{Relationship to Existing Methodologies}

NFD shares conceptual elements with several established approaches while remaining methodologically distinct from each.

Like \textbf{Agile software development}~\cite{boehm1988spiral}, NFD is iterative and values working systems over comprehensive documentation. However, Agile iterations are developer-driven sprints producing code artifacts, while NFD iterations are user-driven conversational cycles producing knowledge artifacts. The ``customer'' in Agile is a stakeholder who provides requirements; the ``user'' in NFD is the domain expert who \emph{is} the primary knowledge source.

Like \textbf{cognitive apprenticeship}~\cite{collins1989cognitive}, NFD emphasizes learning through situated practice. The agent occupies a role analogous to an apprentice: initially possessing general capability but lacking domain-specific judgment, and acquiring expertise through observation of and interaction with an experienced practitioner. The key difference is \emph{bidirectionality}---the agent simultaneously serves as a memory and analytical resource for the practitioner, creating a symbiotic rather than purely pedagogical relationship.

Like \textbf{knowledge management systems}~\cite{nonaka1995knowledge, alavi2001knowledge}, NFD addresses expertise capture and organization. However, traditional KM treats knowledge as an organizational resource to be collected, while NFD treats it as a co-constructed cognitive asset existing in the interaction between a specific human and agent.

Like \textbf{RLHF}~\cite{ouyang2022training}, NFD aligns agent behavior with user preferences. However, RLHF operates at the model weight level through gradient-based optimization, while NFD operates at the knowledge representation level through memory accumulation and crystallization---enriching the context within which the model operates without modifying the model itself.

\subsection{Limitations and Open Challenges}

\textbf{Cold start problem.} An agent in the early stages of nurturing provides limited value, potentially discouraging the sustained engagement required to reach productive capability. Historical data migration partially mitigates this but cannot substitute for genuine conversational nurturing, as migrated data lacks the reasoning traces and contextual richness of interactive exchange.

\textbf{Scalability to organizations.} NFD produces highly personalized agents optimized for individual users. Extending to organizational contexts requires addressing how individually-nurtured agents can share and synthesize knowledge---raising questions about knowledge ownership, conflict resolution, and aggregation of diverse experiential corpora.

\textbf{Quality assurance.} Unlike code (unit-testable) or prompts (benchmarkable), the quality of nurtured knowledge is difficult to assess objectively. The agent may absorb user biases alongside genuine expertise. Developing crystallization quality metrics---analogous to test coverage---is an open research direction.

\textbf{Context window economics.} The Three-Layer Architecture mitigates but does not eliminate finite context constraints. The Constitutional Layer must remain concise (10--15\% of context), and the Experiential Layer's value depends on memory search quality. Advances in long-context models and efficient retrieval will directly expand NFD's ceiling.

\textbf{Crystallization bottleneck.} Crystallization currently requires semi-manual direction. Fully automated crystallization---where agents self-identify patterns and propose consolidation~\cite{gao2025evolver}---remains an open challenge. Periodic self-reflection mechanisms in platforms like OpenClaw~\cite{openclaw2025} provide natural trigger points for self-directed crystallization.

\subsection{Broader Implications}

\textbf{Knowledge assets and transferability.} NFD creates a spectrum of transferable knowledge assets. \emph{Scaffold packages} (workspace templates, skill definitions) are highly portable but carry minimal domain value. \emph{Crystallized knowledge packages} (populated skill references, decision frameworks, case libraries) represent substantive domain knowledge but require de-personalization for transfer. The transferability-value tradeoff is inherent to the paradigm's emphasis on personalized expertise.

\textbf{Reflexive benefits for practitioners.} The act of nurturing an agent functions as a structured reflection practice~\cite{kolb1984experiential}. Externalizing tacit knowledge through dialogue creates a feedback loop where nurturing the agent simultaneously deepens the practitioner's own self-understanding. The case study analyst reported that the NFD process revealed previously unrecognized inconsistencies in their decision-making framework. This bidirectional value creation---agent capability \emph{and} practitioner insight---is a distinctive property that sets NFD apart from purely tool-building approaches.

\textbf{New roles in agent development.} NFD implies the emergence of the \emph{agent nurturer}---a practitioner who develops domain-expert agents through sustained conversational interaction and periodic crystallization, deploying domain expertise as the primary development tool rather than code or prompts.

% ============================================================
% 9. Conclusion and Future Work
% ============================================================
\section{Conclusion}
\label{sec:conclusion}

We have proposed Nurture-First Development as a paradigm for building domain-expert AI agents, addressing the fundamental mismatch between the tacit, evolving nature of domain expertise and the static nature of traditional agent configurations. The core innovation lies in reframing agent development as a \emph{continuous, conversational process}, with the Knowledge Crystallization Cycle as the mechanism that transforms fragmented experiential knowledge into structured, reusable assets. The Three-Layer Cognitive Architecture organizes knowledge by volatility and personalization, while the Dual-Workspace Pattern and Spiral Development Model provide a practical operational framework that accommodates both structural development and conversational growth.

NFD redefines who the ``developer'' of a domain-expert agent is: not the engineer or prompt engineer, but the \emph{domain practitioner} whose expertise is being encoded. The agent's value derives from accumulated experiences, crystallized patterns, and depth of alignment with its human partner---properties that compound over time through sustained use, mirroring the dynamics of expertise itself rather than those of software.

Future work should address: (1)~\emph{automated crystallization algorithms} enabling self-directed knowledge consolidation; (2)~\emph{multi-agent knowledge sharing} for organizational contexts; (3)~\emph{crystallization quality metrics} providing objective signals for knowledge asset quality; (4)~\emph{nurturing interaction design} studying which conversational patterns produce highest-quality experiential data; and (5)~\emph{longitudinal empirical studies} across diverse domains to validate the theoretical framework.

% ============================================================
% References
% ============================================================
\bibliographystyle{unsrtnat}
\bibliography{references}

\end{document}